\documentclass[acmtog, authorversion]{acmart}

\usepackage{booktabs} 

\setcitestyle{square}

\usepackage{subfig}

\usepackage[ruled]{algorithm2e} 

\SetAlFnt{\small}
\SetAlCapFnt{\small}
\SetAlCapNameFnt{\small}
\SetAlCapHSkip{0pt}
\IncMargin{-\parindent}

\setcopyright{none}

\newcommand{\Fref}[1]{Fig.~\ref{#1}}
\newcommand{\Sref}[1]{Sec.~\ref{#1}}

\begin{document}


\title{Comicolorization: Semi-Automatic Manga Colorization}

\author{Chie Furusawa*}
\affiliation{%
\institution{DWANGO Co., Ltd.}
\country{Japan}}

\author{Kazuyuki Hiroshiba*}
\affiliation{%
\institution{DWANGO Co., Ltd.}
\country{Japan}}

\author{Keisuke Ogaki}
\affiliation{%
\institution{DWANGO Co., Ltd.}
\country{Japan}}

\author{Yuri Odagiri}
\affiliation{%
\institution{DWANGO Co., Ltd.}
\country{Japan}}

\begin{abstract}



We developed \textit{Comicolorization}, a semi-automatic colorization system for manga images. Given a monochrome manga and reference images as inputs, our system generates a plausible color version of the manga.
This is the first work to address the colorization of an entire manga title (a set of manga pages).
Our method colorizes a whole page (not a single panel) semi-automatically, 
with the same color for the same character across multiple panels.
To colorize the target character by the color from the reference image, we extract a color feature from the reference and feed it to the colorization network to help the colorization.
Our approach employs adversarial loss to encourage the effect of the color features.
Optionally, our tool allows users to revise the colorization result interactively.
By feeding the color features to our deep colorization network, we accomplish colorization of the entire manga using the desired colors for each panel.
\end{abstract}

%
%


%
%


\begin{teaserfigure}
  \centering
  \includegraphics[height=12.5cm]{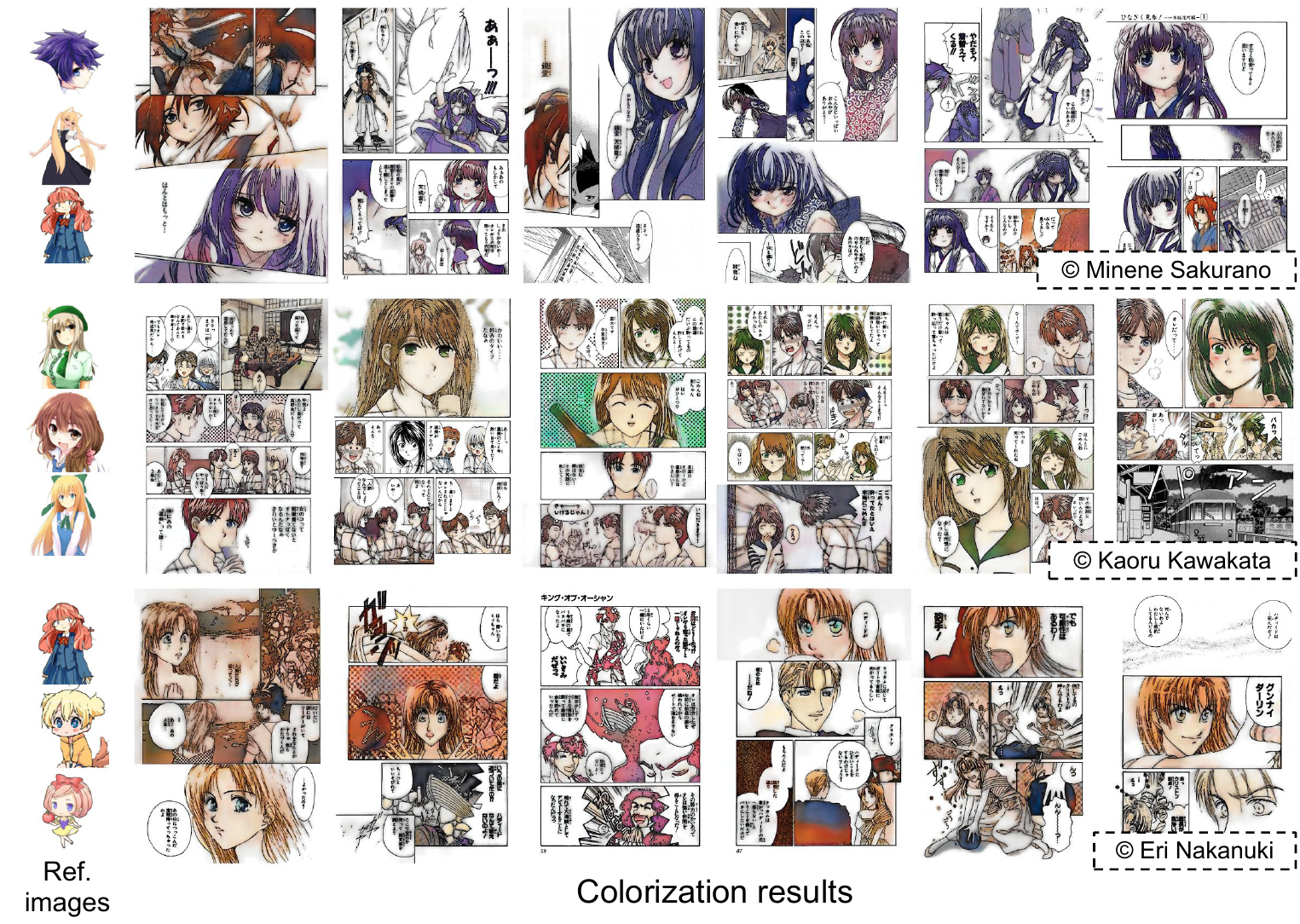}
   \caption{
   	Colorization results by the proposed system.
   	Each row illustrates pages from the same manga title.
   	The leftmost column shows reference images. Each manga page is colorized by the reference images.
	Note that we do not provide manual interactive revisions for these images.
   	All manga are from the Manga 109 dataset \protect\cite{manga109}.}
   \label{fig:teaser}
\end{teaserfigure}

\thanks{*:The first two authors are contributed equally.}




\fancypagestyle{standardpagestyle}{}
\fancypagestyle{firstpagestyle}{}
\fancyfoot{}
\fancyhead{}

{
  \makeatletter
  \g@addto@macro\@subtitlenotes{
    \let\@makefnmark\relax  \let\@thefnmark\relax
    \let\@makefntext\noindent
    \footnotetextcopyrightpermission{
      \def\par{\let\par\@par}\parindent\z@\@setthanks
    }
    \renewcommand\footnotetextcopyrightpermission[1]{}
  }
  \maketitle
  \makeatother
}

\section{Introduction}
Manga (Japanese comics) have attracted readers all over the world.
Because manga are usually created with a pen and ink on white paper, most existing manga are monochrome.
To make the existing manga more attractive, there is a strong demand for a colorization technique for monochrome manga.
However, manual colorization of manga is a time-consuming task because the artist needs to edit each page one by one.

To efficiently colorize manga, full automatic colorization methods with convolutional neural networks (CNNs) have been proposed \cite{DeepColorization,LTBC,pix2pix,ColorfulImageColorization}. 
However, it is impossible for such full automatic methods to beautifully colorize a whole manga title (a set of manga pages) because of the ``color ambiguity problem.''

\begin{figure}[tb]
  \centering
  \includegraphics[width=\linewidth]{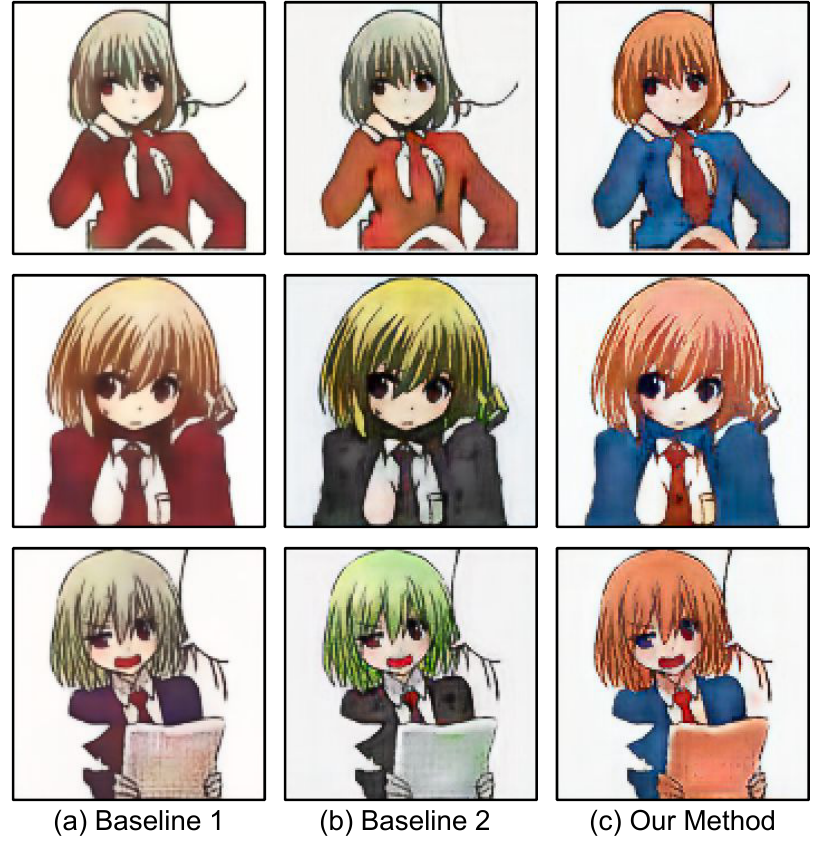}
  \caption{
Examples of the color ambiguity problem: (a) Results by \protect\cite{LTBC}. (b) Results by \protect\cite{LTBC} + adversarial loss \protect\cite{GAN}. (c) Our results.
}
  \label{fig:color_ambiguity}
\end{figure}

The color ambiguity problem is defined as follows.
A manga title typically consists of hundreds of pages, where each page is composed of several panels (\Fref{fig:teaser}).
It is therefore essential to colorize the same character by the same color across all panels.
This is not a trivial task, because (1) the same character might be visually different,
and (2) different characters might seem visually similar.
This is the color ambiguity problem, and is the reason that fully automatic methods do not work well.

\Fref{fig:color_ambiguity} illustrates an example. 
Each column contains three panels with the same character, where the automatic method does not work well.
The first and second columns show the colorized results by fully automatic methods. \Fref{fig:color_ambiguity}(a) shows the results by \cite{LTBC}. \Fref{fig:color_ambiguity}(b) shows the results by \cite{LTBC} + adversarial loss \cite{GAN}.
Our objective is to colorize three panels beautifully by the same color composition.
\Fref{fig:color_ambiguity}(a) and (b) show two typical tendencies of the automatic methods.
\Fref{fig:color_ambiguity}(a) is colorized by ``faint'' colors, i.e., reddish and non-vivid in this case.
By this type of method, all images are colorized by such faint and non-vivid colors.
On the other hand, each character of \Fref{fig:color_ambiguity}(b) is colorized by vivid colors (e.g., the fresh green hair color of the third row). 
However, the same character might be colorized by the different colors, such as the yellow hair of the second row.
These two tendencies caused by the color ambiguity problem are typical of automatic methods. To colorize the same character by the same color, the results become faint and non-vivid.
If the system colorizes characters more vividly, the same character would have the different colors.

\begin{figure*}[tb]
  \centering
  \includegraphics[width=\textwidth]{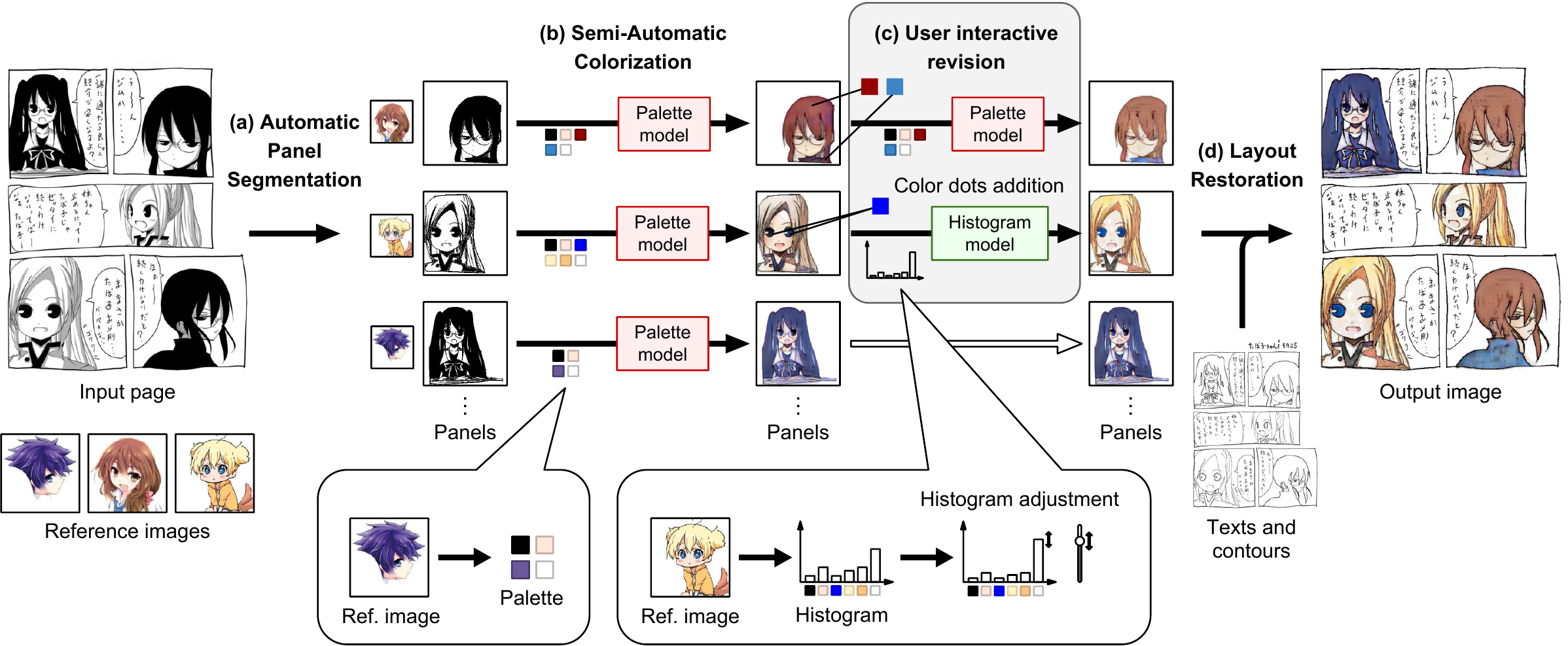}
  \caption{
Illustration of our colorization pipeline. The inputs of the system are manga page images and reference images. (a) The system then segments the manga page images into panel images for colorization (described in \Sref{prepro}). (b) The colorization step is the main operation of our tool. This task is performed semi-automatically using input reference images (described in \Sref{mainpro}). (c) Then, the revision is carried out by the user. The user can interactively modify the result by applying color dots and/or adjust color histogram (described in \Sref{revision}). (d) Finally, the colorized panels are restored to the original layout via resizing and overlaying the text and contours from the original page (described in \Sref{postpro}).
}
  \label{fig:overview}
\end{figure*}

To solve the color ambiguity problem, we propose a semi-automatic colorization method (\Fref{fig:overview}).
To colorize the input manga images, our system takes reference images as additional inputs. 
From each reference image, a color feature is extracted.
Our system colorizes the manga images using the extracted color feature.
\Fref{fig:color_ambiguity}(c) shows the result of the proposed approach with a reference image. 
The characters are colorized not only vividly but also with the same color for the same area, such as orange hair.
Compared to the manual colorization, our system drastically reduces the operations
because only a small number of interactions are required (only selecting reference images for each character).
\Fref{fig:teaser} illustrates other examples, where several manga pages were colorized using reference images.

To further refine the result, our system provides standard interactive refinement functions,
such as providing color dots \cite{CNNphotoWithDot2017,Scribbler} or controlling the dominant color.
A user who is not satisfied with the result can interactively modify it.

\section{Proposed System}
\Fref{fig:overview} shows an overview of our system.
The pre-processing step, including the automatic panel segmentation, is explained in \Sref{prepro}.
\Sref{mainpro} introduces the semi-automatic colorization method.
The revision functions are explained in \Sref{revision}.
\Sref{postpro} describes the post-processing step for obtaining a complete colorized manga page.

\subsection{Automatic Panel Segmentation\label{prepro}}
Our system requires manga pages and reference images of each character as inputs. 
First, we binarize the input manga.
Then, our system segments the input manga pages into panels for the colorization phase. 
Panel detection and segmentation are performed automatically on the basis of \cite{Chad}. 
This process is illustrated in \Fref{fig:overview}(a).

\subsection{Semi-Automatic Colorization\label{mainpro}}
Each segmented panel (character) is colorized using our CNN architecture (\Fref{fig:overview}(b)).
Our architecture is based on \cite{LTBC}, which is a convolutional encoder-decoder network with an additional classification branch.
Our model has three improvements over theirs:
\begin{itemize}
\item leveraging color features from reference images, 
\item training the classification branch using character names,
\item training with adversarial loss for vivid colorization.
\end{itemize}
Please see Sec.5  for the details of our architecture and training procedure. 
All source code is available. \footnote{ \href{https://github.com/DwangoMediaVillage/Comicolorization} {https://github.com/DwangoMediaVillage/Comicolorization} } 

\textbf{Color feature:}
We extract color features from reference images, and feed them into the CNN architecture to achieve
consistent colorization for the same character.
Given a reference image, we scan all pixels in order to create a color histogram.
The histogram is first normalized by the number of pixels, and quantized from a typical $256^3$ bins to $6^3$ bins.
The resultant color histogram for each reference is $\mathbf{h} \in \mathbb{N}^{216}$.
We further binarize this histogram to $\mathbf{p} \in \{0, 1\}^{216}$ to handle the color feature more implicitly.
We call this 216-dimensional binary vector $\mathbf{p}$ a \textit{palette}.
Compared to the color histogram, a palette is robust against the difference of the ratio of colors between the reference image and the input image.

\textbf{Classification training by character names:}
The original model of \cite{LTBC} has a classification branch.
This branch predicts the category of the scene to better illustrate the result.
For example, if the model recognizes an input image as sea,
the model tends to colorize the image using more blue than green.
In our case, we train this categorization branch by the pair of the name of the character and its image, using our character dataset.

\textbf{Adversarial loss for vivid color:}
Additionally, we add an adversarial loss to the objective function to this architecture \cite{GAN,DCGAN}.
Recent studies showed that the adversarial loss helps to colorize images more vividly \cite{pix2pix,Scribbler}, and we also follow this line.
At training time, we input ground-truth and generated images to the discriminator one by one, and the discriminator outputs 0 (= Fake) or 1 (= Real).
Following the previous approaches, the loss of the discriminator is sigmoid cross entropy.

\subsection {User Interactive Revision\label{revision}}
Optionally, our system allows users to interactively revise the colorized results (\Fref{fig:overview}(c)).
Our system has two functions for revision: histogram and color dots.
The histogram is used for global revision of the panel, and color dots are used for local revision.

\textbf{Histogram:}
If the user fails to select colors by a palette, he or
she can interactively change the color feature from the palette $\mathbf{p}$ to its original histogram $\mathbf{h}$.
The user can manually adjust the amount of the most frequent color bin (typically, the background color).
The color feature is then automatically normalized with the adjustment.
This adjustment of the frequent color bin is useful to for preventing color bleeding.

\textbf{Color dots:}
Our method can accept color dots in addition to color features.
If a user applies a color dot to a region, the area around the selected region is colorized by the color of the dot.
Because our model is trained using not only typical images but also images with synthesized color dots,
we can directly feed an input image with color dots to our model.

\subsection{Layout Restoration\label{postpro}}
Finally, we finish the colorization by the following four steps:
To improve the results, whitish and blackish areas are automatically changed to pure black and white, respectively.
To resize the panel images to the original size, we perform super-resolution.
Finally, the colorized panels are re-composed to the original page.
The texts and contours from the original page are then overlaid.
This process is illustrated in \Fref{fig:overview}(d).

\section{Training}
In this section, we briefly explain how to train the network for colorization.

We collected hand-drawn character images as a training dataset instead of manga images
because there are very few pairs of monochrome and colorized manga images publicly available.
We binarized each color manga image using \cite{Otsuthreshold} to make a pair of color and monochrome images.
The original color image is used as an input, and the monochrome image is used as a ground truth for the training process.

Given pairs of training images (input and ground truth), 
our network was trained based on the mean squared error criterion over the L*a*b color space,
with an additional adversarial loss and a classification loss.

To handle revision by color dots, we trained our network using synthesized color dots that are created on the basis of the colors in the ground truth images.

\section{Results and Discussion}
\begin{figure}[tb]
  \begin{center}
  \includegraphics[width=\linewidth]{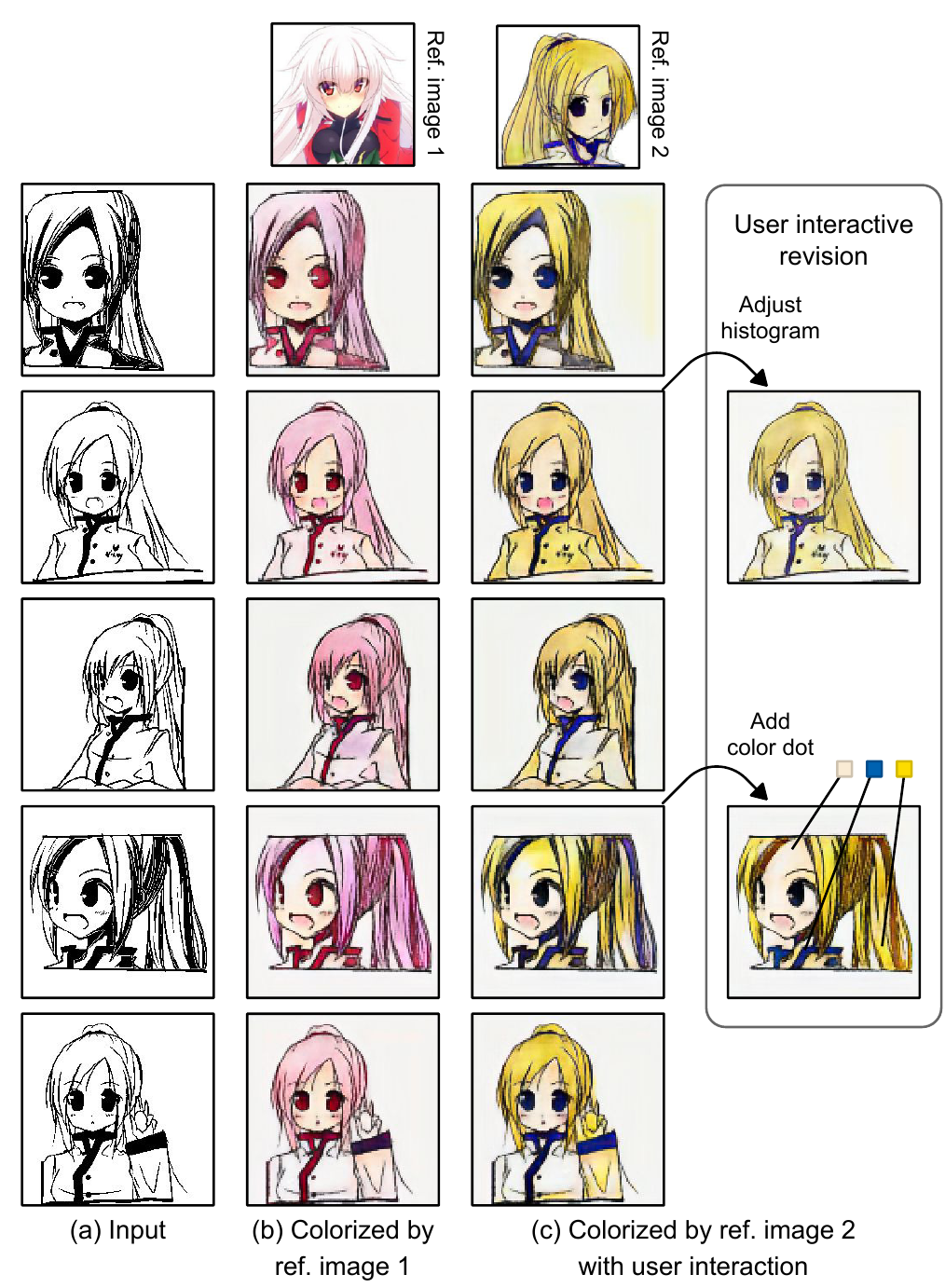}
  \caption{
  	Results generated from our model and others. (a) Image based on \protect\cite{LTBC}. (b), (c) Colorized images using our model with the palette extracted from the left reference image. 
  }
  \label{fig:our_results}
  \end{center}
\end{figure}

\Fref{fig:our_results} shows inputs and the colorization results of our model.
\Fref{fig:our_results}(a) shows the input images.
\Fref{fig:our_results}(b) shows the colorized results using the palette of the reference images at the top.
In the case of \Fref{fig:our_results}(c), the color version of the input character was used as a reference.
Our method successfully colorized a set of input characters vividly with the same color composition, using a reference character.
\Fref{fig:our_results}(c) was revised via manual revisions as shown in the next column. 
The bleeding of the yellow color to the girl's clothes on the left side was prevented by adjusting the histogram.
Incomplete areas such as the forehead, hair, and collar were successfully colorized by providing color dots.
Because characters across different panels can be easily colorized using the same color, our method is highly suitable for manga colorization.

\textbf{Limitations:}
Our semi-automatic colorization algorithm sometimes fails to colorize the panel image when multiple characters appear.
In the case of such failure, the user can edit the image with many color dots.

\begin{figure*}[htbp]
  \centering
  \includegraphics[width=\textwidth]{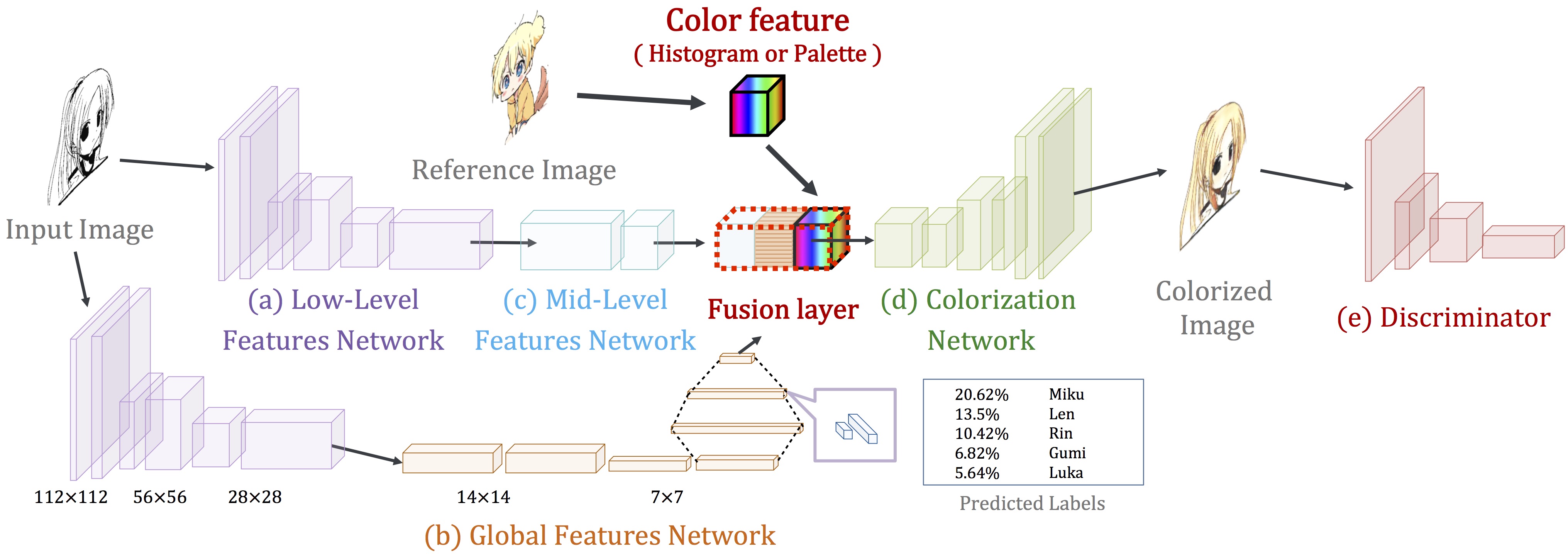}
  \caption{
  	Network architecture at the colorization step.
  }
  \label{fig:network}
\end{figure*}

\section{Colorization Network Model Detail}
\subsection{Training Dataset\label{dataset}}
We crawled several color illustrations from \textit{niconico-seiga} \footnote{ \href{http://seiga.nicovideo.jp/} {http://seiga.nicovideo.jp/} }, which is one of the largest illustration-sharing website in Japan.
In niconico-seiga, creators draw an illustration of a character, and post it on the service with the name of the character.

To train our model, we leveraged the \textit{niconico-seiga} dataset,  which contains these images.
Because our objective is to colorize manga vividly,
we removed images that have a low saturation value in the HSV color space from the dataset.
Because our main target is a character, we used only character images that were detected by an animation-face-detector \footnote{ \href{https://github.com/nagadomi/lbpcascade\_animeface} {https://github.com/nagadomi/lbpcascade\_animeface} }.
We eventually used a total of 160,000 images as the training dataset. 

First, each image was resized to 256$\times$256 pixels.
We then randomly cropped regions to 224$\times$224 pixels from each image.
This cropping makes our model robust.
For further robustness, we also randomly flipped the images horizontally with 50\% probability.

\subsection{Network Architecture}
\Fref{fig:network} shows the architecture of the networks in our system.
This architecture was based on \cite{LTBC}, whose model has four networks:
a global features network, low-level features network, mid-level features network, and colorization network.
In addition to \cite{LTBC}, we added three components:
leveraging color features from reference images, 
training the classification branch using character names,
and training with adversarial loss for vivid colorization.
We explain the implementation details of
(1) fusing the color features into the network, (2) preparing the data for the global network, and (3) setting the parameters.

\textbf{How to fuse color features:} 
In the original model of \cite{LTBC}, 
given an input image, the visual information (the weight tensor in $\mathbb{R}^{28 \times 28 \times 256}$)
is extracted via low- and mid-level feature networks.
The global information (the weight vector in $\mathbb{R}^{256}$ that represents a category of the image)
is extracted via global feature networks.
These two features are concatenated so as to form a latent variable of the network (``fusion layer'' in \Fref{fig:network})
\footnote{
For concatenation, the 256-dimensional global feature is duplicated to $28\times28$ times,
i.e., the ($u, v$) position of the resultant latent variable is the concatenation of 
(1) the ($u, v$) position of the local feature and (2) the global feature.
}.

In our proposed network, we further concatenated a 256-dimensional color feature to the fusion layer
in the same manner as the concatenation of the global feature.
With the color feature, the color information of a reference image is automatically incorporated to colorize the input image.

\textbf{Data preparation for the global network:}
The global features network of the original model predicts the label (scene category) of the input image.
Unlike the original model, the global network of our model predicts the label of the name of the character.
The label information was based on the service of \textit{niconico-seiga} (as shown \Sref{dataset}).
The total number of collected labels was 428.

\textbf{Parameter settings:} 
Table 1 illustrates the detailed parameter settings of the networks.
We trained the model for 550,000 iterations with a batch size of 30.

As an activation function, we used a sigmoid function for the final output layer, and ReLU for the other layers.
We inserted a batch-normalization layer \cite{BN} after each convolution layer. 
\captionsetup[subfloat]{position=top}

\begin{table*}[tb]
\centering
\subfloat[][Low-Level Features network]{
\begin{tabular}{cccc} \hline
        \multicolumn{1}{c}{Type}
        & \multicolumn{1}{c}{Kernel} & \multicolumn{1}{c}{Stride} & \multicolumn{1}{c}{Outputs} \\ \hline \hline
        conv. & 3$\times$3 & 2$\times$2 & 64 \\ 
        conv. & 3$\times$3 & 1$\times$1 & 128 \\ \hline
        
        conv. & 3$\times$3 & 2$\times$2 & 128 \\ 
        conv. & 3$\times$3 & 1$\times$1 & 256 \\ \hline

        conv. & 3$\times$3 & 2$\times$2 & 256 \\
        conv. & 3$\times$3 & 1$\times$1 & 512 \\ \hline
\end{tabular}
}
\hspace{20pt}
\subfloat[][Global Features network]{
      \begin{tabular}{cccc} \hline
        \multicolumn{1}{c}{Type}
        & \multicolumn{1}{c}{Kernel} & \multicolumn{1}{c}{Stride} & \multicolumn{1}{c}{Outputs} \\ \hline \hline
        conv. & 3$\times$3 & 2$\times$2 & 512 \\ 
        conv. & 3$\times$3 & 1$\times$1 & 512 \\ \hline
        conv. & 3$\times$3 & 2$\times$2 & 512 \\ 
        conv. & 3$\times$3 & 1$\times$1 & 512 \\ \hline
        FC & - & - & 1024 \\
        FC & - & - & 512 \\ \hline
\end{tabular}
}
\hspace{20pt}
\subfloat[][Mid-Level Features network]{
        \begin{tabular}{cccc} \hline
          \multicolumn{1}{c}{Type}
          & \multicolumn{1}{c}{Kernel} & \multicolumn{1}{c}{Stride} & \multicolumn{1}{c}{Outputs} \\ \hline \hline
          conv. & 3$\times$3 & 1$\times$1 & 512 \\ 
          conv. & 3$\times$3 & 1$\times$1 & 256 \\ \hline
\end{tabular}
}

\subfloat[][Colorization network]{
      \begin{tabular}{cccc} \hline
        \multicolumn{1}{c}{Type}
        & \multicolumn{1}{c}{Kernel} & \multicolumn{1}{c}{Stride} & \multicolumn{1}{c}{Outputs} \\ \hline \hline
        
        fusion & - & - & 256 \\ 
        conv. & 3$\times$3 & 1$\times$1 & 128 \\ \hline
        
        upsample & - & - & 128 \\ 
        conv. & 3$\times$3 & 1$\times$1 & 64 \\
        conv. & 3$\times$3 & 1$\times$1 & 64 \\ \hline
        
        upsample & - & - & 64 \\ 
        conv. & 3$\times$3 & 1$\times$1 & 32 \\
        conv. & 3$\times$3 & 1$\times$1 & 3 \\ \hline
\end{tabular}
}
\hspace{20pt}
\subfloat[][Discriminator]{
      \begin{tabular}{cccc} \hline
        
      \multicolumn{1}{c}{Type}
      & \multicolumn{1}{c}{Kernel} & \multicolumn{1}{c}{Stride} & \multicolumn{1}{c}{Outputs} \\ \hline \hline
      
      conv. & 4$\times$4 & 2$\times$2 & 64 \\ \hline
      conv. & 4$\times$4 & 2$\times$2 & 128 \\ \hline
      
      conv. & 4$\times$4 & 2$\times$2 & 256 \\ \hline
      conv. & 4$\times$4 & 2$\times$2 & 512 \\ \hline
\end{tabular}
}
\caption{Parameters of our networks}
\end{table*}

\subsection{Optimization}
Our network was trained on the basis of the mean squared error (MSE) criterion over the L*a*b color space,
with an additional adversarial loss and a classification loss.
These losses were weighted as:
\begin{equation}
 MSE:adversarial:classification = 1:1:0.003.
\end{equation}

We trained our network using the Adam optimizer~\cite{Adam}.	
We used the Adam parameters $\alpha$ = 0.0001, $\beta_1$ = 0.9, and $\beta_2$ = 0.999. 
We used the optimization procedure until convergence of the loss.

\subsection{Generation of Training Color Dots\label{colordots}}
Our system enables users to interactively revise the colorized result using color dots.
To train a network to recognize these control color dots at test time,
we synthesized pseudo color dots for the training data.
We generate synthesized color dots on the basis of the colors in the ground truth image.

We put synthesized color dots at random positions on the input monochrome image.
Each pixel of the resultant image consists of three values.
The first value is zero or one, which represents the pixel value of the original input monochrome image.
The second and the third values are the ``a'' and ``b'' values, respectively, of the color dots in the L*a*b color space.
If the color dot does not exist in the pixel, zero is padded for the second and the third value.

The size of synthesized color dots is one pixel, 
and the number of the dots is decided randomly from 0 to 15.
The model becomes robust by feeding a random number of dots. 
Our method works perfectly even if the number of the inputs is more than 15, the maximum of the number of given synthesized color dots, except for unusual cases such as all points placed in the same position.
In the testing phase, users can provide any number of points to the system to enhance the result. 

\section{Conclusions}
In this paper, we propose the first system to colorize an entire manga at once. Our method colorizes a whole page (not a single panel).
Moreover, with color features extracted from the input reference image, we confirmed that 
our system was able to colorize the same character by the same color.
By employing adversarial loss, we confirmed that adversarial networks emphasize the effect of given color information. 
We demonstrated that the color features help the inference by our deep colorization network.
In addition, our tool allows users to revise the colorization result interactively by adjusting color histogram and adding color dots.

\section*{APPENDIX}
\appendix
\begin{figure}[t]
  \centering
  \includegraphics[width=\linewidth]{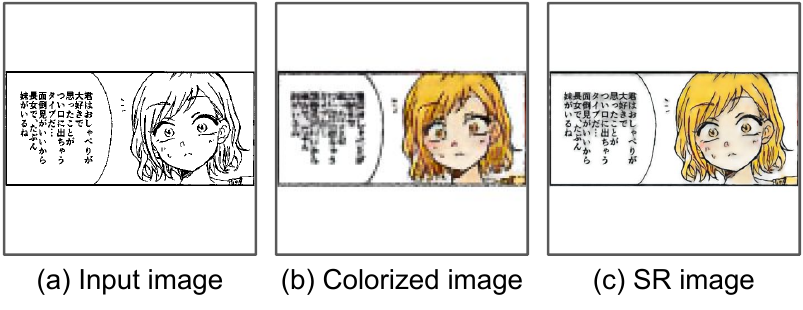}
  \caption{Results after super-resolution. (a) Input monochrome manga panel. (b) Image colorized by our colorization network. (c) Image generated via super-resolution.}
  \label{fig:super_resolution}
\end{figure}

\section{Super Resolution Model}
Although we used character illustrations as the training datasets, 
manga contains special components drawn with lines, such as many contours of panels, text balloons, and the text itself.
Because the resolution of the image is changed during the colorization process,
the contours and texts tend to be blurred unintentionally.
To remove such unintentional blurring, we apply a super-resolution (SR) method for clear colorization.

A sample result of the process is shown in \Fref{fig:super_resolution}.
The architecture of super-resolution is inspired by \textit{PaintsChainer} \footnote{ \href{https://github.com/pfnet/PaintsChainer} {https://github.com/pfnet/PaintsChainer} }, whose super-resolution model was originally based on \cite{U-Net}.
The original size of colorized images by our colorization network is 
$224 \times 224$ pixels for each panel (Fig.\ref{fig:super_resolution} (b)).
By applying super-resolution, we obtain a high-resolution image with 448$\times$448 pixels (\Fref{fig:super_resolution} (c)).

\begin{figure}[t]
  \centering
  \includegraphics[width=14cm,bb=0 0 700 300]{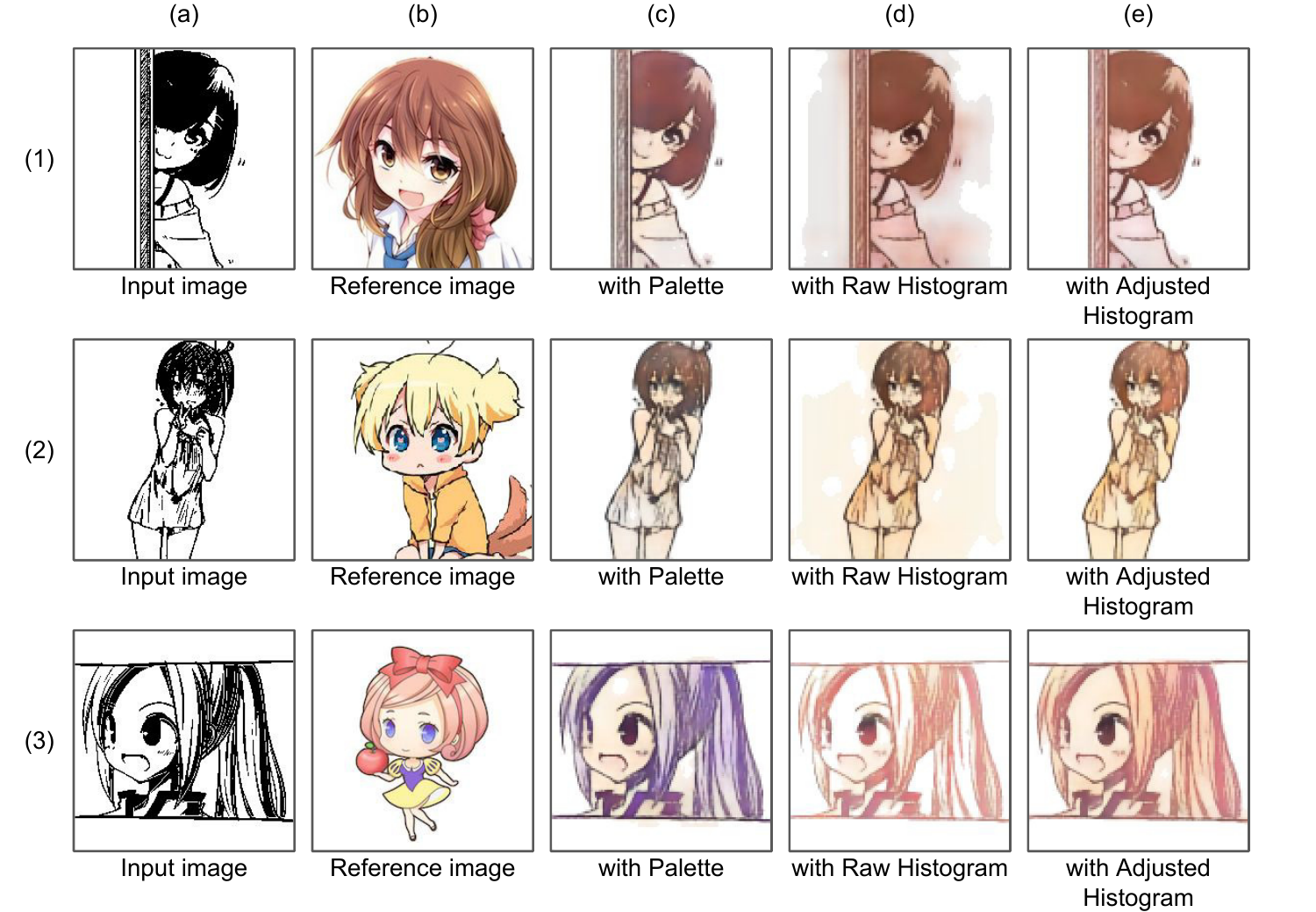}
  \caption{Results of in the case of using a palette, a raw histogram, and an adjusted histogram. (a) Input image, (b) reference image for the histogram or the palette, (c) image generated by the model with the palette, (d) image generated by the model with the raw histogram, and (e) image generated by the model with the adjusted histogram.
  }
  \label{fig:histogram_adjustment}
\end{figure}

\begin{figure}[t]
  \centering
  \includegraphics[width=11cm,bb=0 0 500 150]{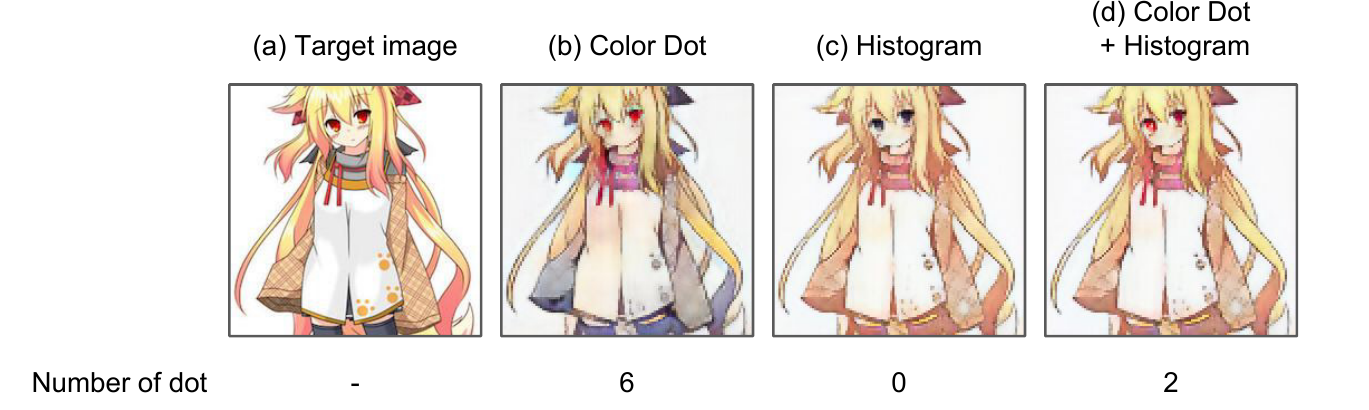}
  \caption{Results of using only color dots, using only histogram, and combining color dots and histogram, as well as the number of color dots used. (a) Target image, and (b) image colorized using six color dots aiming to obtain a similar colorization result as (a). (c) Image colorized using the histogram calculated from the target image (a), and (d) image generated by adding two color dots to (c) and inputting the histogram of (a).
}
  \label{fig:color_dot}
\end{figure}

\section{Interactive Revision}
In this section, we discuss the effect of histogram adjustment and using color dots with results.

\Fref{fig:histogram_adjustment} explains the comparison of color features.
\Fref{fig:histogram_adjustment} shows the results of using a palette, a raw histogram, and an adjusted histogram.
For all rows, the results with a palette (\Fref{fig:histogram_adjustment} (c))
have less color breeding than those with a raw histogram (\Fref{fig:histogram_adjustment} (d)).
This indicates that the palette enables us to robustly colorize images with less breeding, 
because the palette constrains the possible color space.
In the result using a raw histogram (\Fref{fig:histogram_adjustment} (d)),
the dominant color is spread.
At the top-row image, the red-brown color of the hair in the reference image bleeds
beyond the boundaries (Fig. \ref{fig:histogram_adjustment} (1-d));
this occurs in the second row as well (\Fref{fig:histogram_adjustment} (2-d)).
In the third row (\Fref{fig:histogram_adjustment} (3-d)),
the background color is white, and the most frequent color in the reference image is white.
Consequently, the result with a raw histogram is too whitish.
By adjusting the ratio of the dominant color (white), 
the result is improved (\Fref{fig:histogram_adjustment} (e)), where the hair and skin is colorized with pink.

In addition, we introduce how users can combine color features and color dots effectively.
Figure \ref{fig:color_dot} shows the colorized results along with the number of color dots.
The images show the use of color dots (Fig. \ref{fig:color_dot} (b)),
a histogram (Fig. \ref{fig:color_dot} (c)), or a combination of both (\Fref{fig:color_dot} (d)).
When the user adds six dots, the colorization result is as shown in Fig. \ref{fig:color_dot} (b).
The user adds four dots for the hairs (two dots near the head and the others at the hair ends of the right side and the left side),
and two dots to both eyes. 
Although the colorized result is similar to the target (\Fref{fig:color_dot} (a)), several interactions are required (providing six dots).
If the user uses only the histogram extracted from the target (\Fref{fig:color_dot} (a)) as reference,
the colorization result is as shown in \Fref{fig:color_dot} (c). In this image, the eye color differs from that of the target image,
but most regions are colorized using the same colors as the target image.
\Fref{fig:color_dot} (d) shows the result with two additional dots to \Fref{fig:color_dot} (c).
Two dots are added to both eyes. 
This result is colorized like the target by adding fewer color dots in comparison (\Fref{fig:color_dot} (b)).
We recommend that the user colorizes using the color features calculated
from the reference image first and then specifying the local regions by adding only a few color dots.

\section{Histogram Blending}
\begin{figure}[t]
  \centering
  \includegraphics[width=13cm,bb=0 0 500 100]{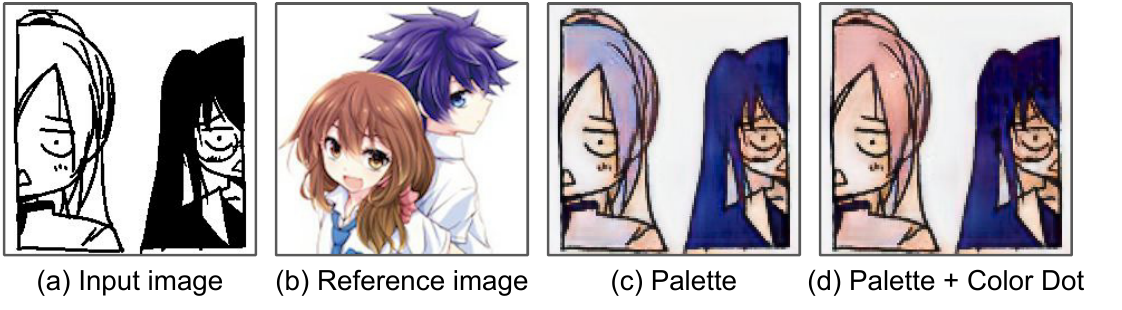}
  \caption{One of the failure cases of our colorization method. (a) Input image where two girls appear in the panel. (b) Reference image of the colorization where one girl has brown hair and one boy has purple hair. (c) Image colorized with the palette of (b). (d) Image revised by the user, who added color dots to the left girl.}
  \label{fig:limitation}
\end{figure}

Our semi-automatic colorization algorithm sometimes fails to colorize the panel image when multiple characters appear (\Fref{fig:limitation}).
In \Fref{fig:limitation} (a), two characters appear in the input panel.
In \Fref{fig:limitation} (c), the girl on the left side is colorized unappealingly, because the hair is colorized using the same color as her hair on the right side. 
As in one of the revision examples, the user edits the image with many color dots (\Fref{fig:limitation} (d)).
However, this interaction wastes the color information from the reference image.

\begin{figure}[t]
  \centering
  \includegraphics[width=13cm,bb=0 0 500 100]{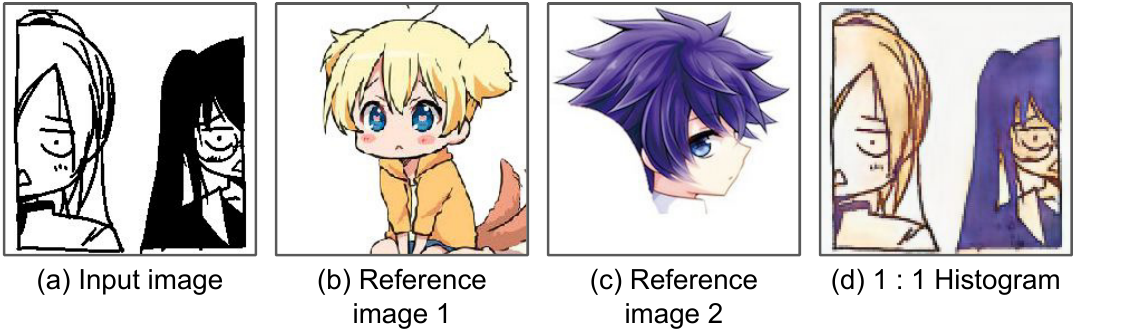}
  \caption{Examples of histogram blending. The image in (d) was generated from (a) with the blend of two histograms. The blended histogram was calculated using one histogram of reference image (b) and another histogram of reference image (c), which was then blended at an equal ratio.}
  \label{fig:histogram_blending}
\end{figure}

Regarding this problem, we consider that histogram blending (\Fref{fig:histogram_blending}) is perhaps suitable for overcoming this problem. 
Instead of using many color dots, we used two reference images (\Fref{fig:histogram_blending} (b) and (c)) at an equal ratio.
In \Fref{fig:histogram_blending} (d), we finally get the suitable colorized result by using the blended histogram.
When the size of each character is different, the user should adjust the blending ratio of the reference images. 

\section{Further Results}
\begin{figure*}[htbp]
  \centering
  \includegraphics[width=\textwidth]{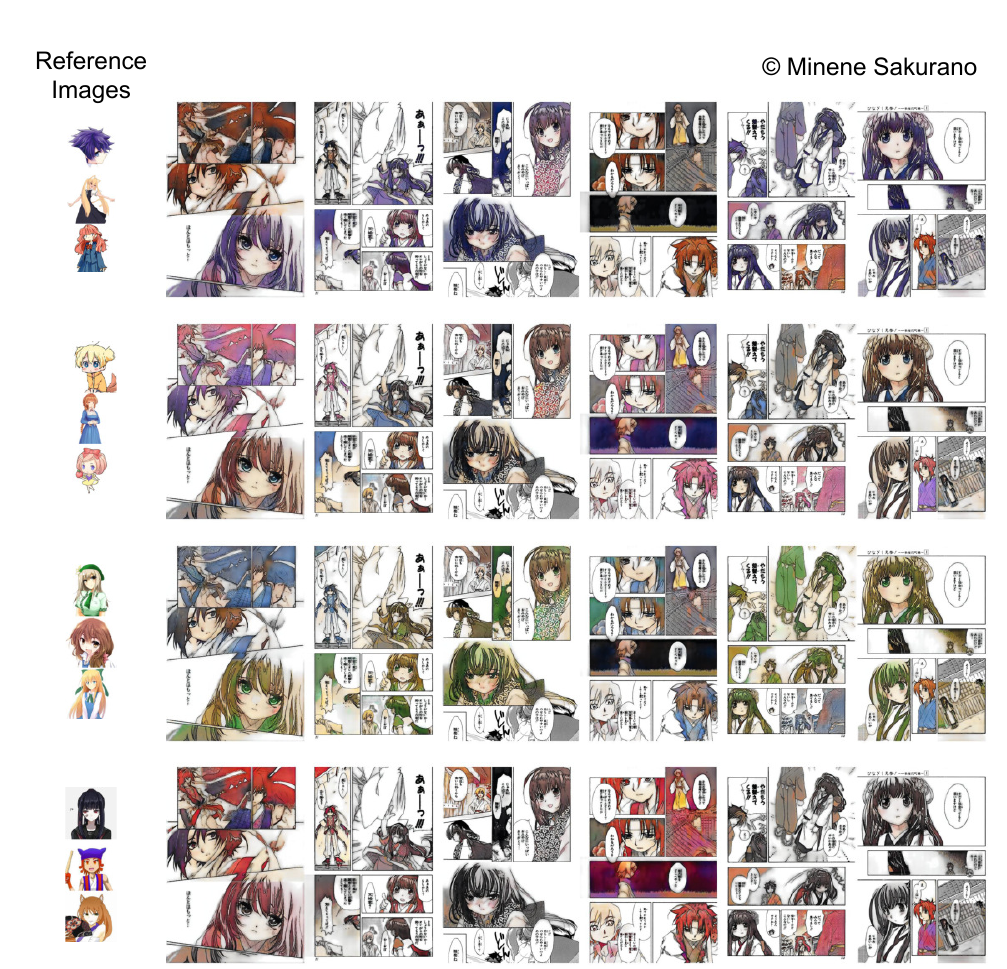}
  \caption{Manga page colorized by our tool. Each manga was colorized semi-automatically using the reference images at left. All input manga were the same. The reference images were different. We used monochrome manga from the Manga 109 dataset \protect\cite{manga109}.}
  \label{fig:further_same_ref}
\end{figure*}

We show further colorization results in \Fref{fig:further_same_ref}.
In \Fref{fig:further_same_ref}, the same manga pages were colorized with different reference images, and reference images were used as the palettes.
Each result was colorized with the colors of each combination of reference images.
 
In addition to this, we colorize all monochrome manga in the Manga 109 Dataset \cite{manga109}
using the palette of the cover image of its manga.

\begin{acks}
We would like to thank all of the illustrators for their inspiring artwork, which has motivated us for this research, and was essential for creating the dataset. 
We would also like to thank Yusuke Matsui and Tsukasa Omoto for the valuable suggestions and fruitful discussions.
\end{acks}

\newpage

\newpage
\newpage

\bibliographystyle{ACM-Reference-Format}
\bibliography{comicolorizatrion-bibliography}

\end{document}